\documentclass[11pt,a4paper]{article}
\usepackage[hyperref]{emnlp-ijcnlp-2019}
\usepackage{times}
\usepackage{latexsym}

\usepackage{multirow}
\usepackage{tabularx}
\usepackage{booktabs}
\usepackage{mathtools}
\usepackage{amssymb}
\usepackage{amsmath}
\usepackage{amsfonts}
\usepackage{bm}
\usepackage{bbm}
\usepackage{pifont}
\usepackage{caption}
\usepackage{subcaption}
\usepackage{adjustbox}

\usepackage{url}
\usepackage{todonotes}
\usepackage{tabto}

\aclfinalcopy 

\newcommand{\cmark}{\textcolor{green}{\ding{51}}}
\newcommand{\xmark}{\textcolor{red}{\ding{55}}}

\DeclareMathOperator*{\argmax}{arg\,max}

\newcommand\copa[3]{
\begin{itemize}
    \item \textbf{Premise}: #1
    \item \textbf{Alternative 1}: #2
    \item \textbf{Alternative 2}: #3
\end{itemize}
}

\newcommand\copafig[3]{
\begin{itemize}
    \item[] {} #1
    \item[] {} #2
    \item[] {} #3
\end{itemize}
}

\title{When Choosing Plausible Alternatives, Clever Hans can be Clever}
\author{Pride Kavumba$^{1,}\thanks{\enspace Equal contribution.}$ \hspace{.9cm} Naoya Inoue$^{1,2,\ast}$ \hspace{.9cm}
Benjamin Heinzerling$^{2, 1}$ \hspace{.9cm}
Keshav Singh$^{1}$\\ 
\bf{Paul Reisert}$^{2,1}$ \hspace{.9cm} {Kentaro Inui$^{1,2}$} \\
 $^1$Tohoku University
 $^2$RIKEN Center for Advanced Intelligence Project (AIP) \\
 \texttt{\{pkavumba, naoya-i, keshav.singh29, inui\} @ecei.tohoku.ac.jp } \\
 \texttt{\{benjamin.heinzerling, paul.reisert\} @riken.jp } \\}

\date{}

\begin{document}
\maketitle
\begin{abstract}
Pretrained language models, such as BERT and RoBERTa, have shown large improvements in the commonsense reasoning benchmark COPA. However, recent work found that many improvements in benchmarks of natural language understanding are not due to models learning the task, but due to their increasing ability to exploit superficial cues, such as tokens that occur more often in the correct answer than the wrong one. Are BERT's and RoBERTa's good performance on COPA also caused by this?
We find superficial cues in COPA, as well as evidence that BERT exploits these cues.
To remedy this problem, we introduce Balanced COPA, an extension of COPA that does not suffer from easy-to-exploit single token cues.
We analyze BERT's and RoBERTa's performance on original and Balanced COPA, finding that BERT relies on superficial cues when they are present, but still achieves comparable performance once they are made ineffective, suggesting that BERT learns the task to a certain degree when forced to. In contrast, RoBERTa does not appear to rely on superficial cues.
\end{abstract}

\section{Introduction}

Pretrained language models such as ELMo \cite{peters2018deep}, BERT \cite{devlin-etal-2019-bert}, and RoBERTa \cite{RoBERTa2019} have led to improved performance in benchmarks of natural language understanding, in tasks such as natural language inference \cite[NLI,][]{liu2019multi}, argumentation \cite{niven2019probing}, and commonsense reasoning \cite{li-etal-2019-learning,socialiqapaper}.
However, recent work has identified superficial cues in benchmark datasets which are predictive of the correct answer, such as token distributions and lexical overlap.
Once these cues are neutralized, models perform poorly, suggesting that their good performance is an instance of the Clever Hans effect\footnote{%
Named after the eponymous horse which appeared to be capable of simple mental tasks but actually relied on cues given involuntarily by its handler.%
} \cite{pfungst1911clever}: Models trained on datasets with superficial cues learn heuristics for exploiting these cues, but do not develop any deeper understanding of the task.

\newlength{\alternativeindent}
\setlength{\alternativeindent}{1.29em}
\definecolor{forestgreen}{HTML}{228B22}
\definecolor{borange}{HTML}{cc5500}
\begin{figure}[t]
    \centering
     \begin{subfigure}[b]{\linewidth}
        \copafig
        {The woman hummed to herself. What was the \emph{cause} for this?}
        {\cmark \tabto{\alternativeindent} She was in \colorbox{orange}
        {a} good mood.}
        {\xmark \tabto{\alternativeindent} She was nervous.}
         \caption{Original COPA instance.}
         \label{fig:original-copa}
     \end{subfigure}
     \par\bigskip 
     \begin{subfigure}[b]{\linewidth}
        \copafig
        {The woman trembled. What was the \emph{cause} for this?}
        {\xmark \tabto{\alternativeindent} She was in \colorbox{orange}{a} good mood.}
        {\cmark \tabto{\alternativeindent} She was nervous.}
         
        \caption{Mirrored COPA instance.}
        \label{fig:mirrored-copa}
     \end{subfigure}
     \hfill
    
    \caption{A COPA instance (a) with premise and correct (\cmark) and wrong (\xmark) alternatives. Our analysis reveals that the unigram \emph{a} (highlighted orange) is a superficial cue exploited by BERT. We neutralize such superficial cues by creating a mirrored instance (b). After mirroring, the highlighted superficial cue becomes ineffective in predicting the correct answer, since it occurs with equal probability in correct and wrong alternatives.}
    \label{fig:balanced-copa}
\end{figure}

While superficial cues have been identified in, among others, datasets for NLI \cite{gururangan-etal-2018-annotation,mccoy2019right}, machine reading comprehension \cite{sugawara-etal-2018-makes}, and argumentation \cite{niven2019probing}, one of the main benchmarks for commonsense reasoning, namely the Choice of Plausible Alternatives \cite[COPA,][]{roemmele2011choice}, has not been analyzed so far.
Here we present an analysis of superficial cues in COPA.

Given a premise, such as \emph{The man broke his toe}, COPA requires choosing the more plausible, causally related alternative, in this case either: because \emph{He got a hole in his sock} (wrong) or because \emph{He dropped a hammer on his foot} (correct).
To test whether COPA contains superficial cues, we conduct a dataset ablation in which we provide only partial input to the model.
Specifically, we provide only the two alternatives, but not the premise, which makes solving the task impossible and hence should reduce the model to random performance.
However, we observe that a model trained only on alternatives performs considerably better than random chance and trace this result to an unbalanced distribution of tokens between correct and wrong alternatives.
Further analysis (\S\ref{sec:easyhardeval}) reveals that finetuned BERT~\cite{devlin-etal-2019-bert}  perform very well (83.9 percent accuracy) on \emph{easy} instances containing superficial cues, but worse (71.9 percent) on \emph{hard} instances without such simple cues.

To prevent models from exploiting superficial cues in COPA, we introduce \emph{Balanced COPA}.
Balanced COPA contains one additional, \emph{mirrored} instance for each original training instance.
This mirrored instance uses the same alternatives as the corresponding original instance, but introduces a new premise which matches the \emph{wrong} alternative of the original instance, e.g. \emph{The man hid his feet}, for which the correct alternative is now because \emph{He got a hole in his sock} (See another example in Figure~\ref{fig:balanced-copa}).
Since each alternative occurs exactly once as correct answer and exactly once as wrong answer in Balanced COPA, the lexical distribution between correct and wrong answers is perfectly balanced, i.e., superficial cues in the original alternatives have become uninformative.

Balanced COPA allows us to study the impact of the presence or absence of superficial cues on model performance.

Since BERT exploits cues in the original COPA, we expected performance to degrade when training on Balanced COPA. However, BERT trained on Balanced COPA performed comparably overall.
As we will show, this is due to better performance on the ``hard'' instances. This suggests that once superficial cues are made uninformative, BERT learns the task to a certain degree.  

In summary, our contributions are:
\begin{itemize}
    \item We identify superficial cues in COPA that allow models to use simple heuristics instead of learning the task (\S\ref{sec:cues});
    \item We introduce Balanced COPA, which prevents models from exploiting these cues (\S\ref{sec:balanced-copa});
    \item Comparing models on original and Balanced COPA, we find that BERT heavily exploits cues when they are present, but is also able to learn the task when they are not (\S\ref{sec:eval}); and
    \item We show that RoBERTa does not appear to exploit superficial cues.
\end{itemize}

\section{Superficial Cues in COPA}
\label{sec:cues}

 \subsection{COPA: Choice of Plausible Alternatives}
Causal reasoning is an important prerequisite for natural language understanding.
The Choice Of Plausible Alternatives (COPA)~\cite{roemmele2011choice} is dataset that aims to benchmark causal reasoning in a simple binary classification setting.\footnote{\url{http://people.ict.usc.edu/~gordon/copa.html}}
COPA requires classifying sentence pairs consisting of the first sentence, the \emph{premise}, and a second sentence that is either cause of, effect of, or unrelated to premise.
Given the premise and two \emph{alternatives}, one of which has a causal relation to the premise, while the other does not, models need to choose the more plausible alternative.
Figure~\ref{fig:original-copa} shows an example of a COPA instance.
The overall 1000 instances are split into training set\footnote{This set is called \emph{development set} by \citet{roemmele2011choice}, but is used as training set by supervised models.} and test set of 500 instances each.

Prior to neural network approaches, the most dominant way of solving COPA was via Pointwise Mutual Information (PMI)-based statistics using a large background corpus between the content words in the premise and the alternatives~\cite{gordon_commonsense_2011-1, Luo:2016:CCR:3032027.3032078, sasaki-etal-2017-handling, goodwin-etal-2012-utdhlt}.
Recent studies show that BERT and RoBERTa achieve considerable improvements on COPA (see Table~\ref{tab:copaboom}).

However, recent work found that the strong performance of BERT and other deep neural models in benchmarks of natural language understanding can be partly or in some cases entirely explained by their capability to exploit superficial cues present in benchmark datasets.
For example, \citet{niven2019probing} found that BERT exploits superficial cues, namely the occurrence of certain tokens such as \emph{not}, in the Argument Reasoning Comprehension Task~\cite{habernal-etal-2018-argument}.
Similarly, \citet{gururangan-etal-2018-annotation, poliak-etal-2018-hypothesis, Dasgupta2018EvaluatingCI} showed that a simple text categorization model can perform well on the Stanford Natural Language Inference dataset~\cite{snli:emnlp2015} and MultiNLI~\cite{williams-etal-2018-broad} when given incomplete input, even though the task should not be solvable without the full input.
This suggests that the partial input contains unintended superficial cues that allow the models to take shortcuts without learning the actual task.
\citet{sugawara-etal-2018-makes} investigated superficial cues that make questions easier across recent machine reading comprehension datasets. Given the fact that superficial cues were found in benchmark datasets for a wide variety of natural language understanding task, does COPA contain such cues, as well?

\begin{table}[t]
\centering
\adjustbox{max width=\linewidth}{
\begin{tabular}{ll}
\toprule
Model & Accuracy \\
\midrule
BigramPMI~\cite{goodwin-etal-2012-utdhlt} &   63.4	\\
PMI~\cite{gordon_commonsense_2011-1} & 65.4	\\
PMI+Connectives~\cite{Luo:2016:CCR:3032027.3032078} & 70.2 \\
PMI+Con.+Phrase~\cite{sasaki-etal-2017-handling} &	71.4	\\
\midrule
BERT-large~\cite{wang2019superglue} & 70.5 \\
BERT-large~\cite{socialiqapaper} & 75.0 \\
BERT-large~\cite{li-etal-2019-learning} & 75.4 \\
RoBERTa-large (finetuned)\footnote{\url{https://super.gluebenchmark.com/leaderboard}} & 90.6 \\
\midrule
BERT-large (finetuned)*  & 76.5 $\pm$ 2.7 \\
RoBERTa-large (finetuned)* & 87.7 $\pm$ 0.9 \\
\bottomrule
\end{tabular}
}
\caption{Reported results on COPA. With the exception of \cite{wang2019superglue}, BERT-large and RoBERTa-large yields substantial improvements over prior approaches.
See \S\ref{sec:cues} for model details.
* indicates our replication experiments.
}
\label{tab:copaboom}
\end{table}

\subsection{Token Distribution}

One of the simplest types of superficial cues are unbalanced token distributions, i.e tokens appearing more often or less frequently with one particular instance label than with other labels.
For example, \citet{niven2019probing} found that the token \emph{not} occurs more often in one type of instance an argumentation dataset.

Similarly we identify superficial cues --- in this case a single token that appears more frequently in correct alternatives or wrong alternatives --- in the COPA training set.
To find superficial cues in the form of predictive tokens, we use the following measures, defined by \citet{niven2019probing}. Let $\mathbb{T}_{j}^{(i)}$ be the set of tokens in the alternatives for data point ${i}$ with label ${j}$.
The \emph{applicability} ${\alpha}_{k}$ of a token $k$ counts how often this token occurs in an alternative with one label, but not the other:
$$\alpha_{k}=\sum_{i=1}^{n} \mathbbm{1}\left[\exists j, k \in \mathbb{T}_{j}^{(i)} \wedge k \notin \mathbb{T}_{\neg j}^{(i)}\right]$$
The \emph{productivity} $\pi_k$ of a token is the proportion of applicable instances for which it predicts the correct answer:
$$\pi_{k}=\frac{\sum_{i=1}^{n} \mathbbm{1}\left[\exists j, k \in \mathbb{T}_{j}^{(i)} \wedge k \notin \mathbb{T}_{\neg j}^{(i)} \wedge y_{i}=j\right]}{\alpha_{k}}$$
Finally, the \emph{coverage} $\xi_{k}$ of a token is the proportion of applicable instances among all instances: $$ \xi_{k}=\frac{\alpha_{k}}{n}$$

Table~\ref{tab:copa-artifacts} shows the five tokens with highest coverage.
For example, \emph{a} is the token with the highest coverage and appears in either a correct alternative or wrong alternative in 21.2\% of COPA training instances. Its productivity of 57.5\% expresses that it appears in in correct alternatives 7.5\% more often than expected by random chance.
This suggests that a model could rely on such unbalanced distributions of tokens to predict answers based only on alternatives without understanding the task.




To test this hypothesis, we perform a dataset ablation, providing only the two alternatives as input to RoBERTa, but not the premise, following similar ablations by \citet{gururangan-etal-2018-annotation,niven2019probing}.
RoBERTa trained\footnote{See \S\ref{sec:bertforcopa} for experimental setup.} in this setting, i.e. on alternatives only, achieves a mean accuracy of 59.6 ($\pm$ 2.3).
This is problematic because COPA is designed as a choice between alternatives given the premise. Without a premise given, model performance should not exceed random chance. Consequently, a result better than random chance shows that the dataset allows solving the task in a way that was not intended by its creators.
To fix this problem, we create a balanced version of COPA that does not suffer from unbalanced token distributions in correct and wrong alternatives.

\section{Balanced COPA (B-COPA)}
\label{sec:balanced-copa}

To allow evaluating models on a benchmark without superficial cues, we need to make them ineffective.
Our approach is to balance the token distributions in correct alternatives and wrong alternatives in the training set.
Without unbalanced token distributions, we hope models are able to learn other patterns more closely related to the task, e.g. a pair of causally related events, rather than superficial cues.


\begin{table}[t]
    \centering
    \begin{tabular}{ccccc}
    \toprule
        Cue &	App. &	Prod.	& Cov. \\
    \midrule
        in	&  47	 &  55.3	&   9.40 \\
  	    was	&  55	 &  61.8	&   11.0 \\
  	    to	&  82	 &  40.2   &	16.4 \\
  	    the	&  85	 &  38.8   &	17.0 \\
  	    a	&  106	 &  57.5   &	21.2 \\
    \bottomrule
    \end{tabular}
    \caption{Applicability (App.), Productivity (Prod.) and Coverage (Cov.) of the various words in the \emph{alternatives} of the COPA dev set.}
    \label{tab:copa-artifacts}
\end{table}

\subsection{Data Collection}

To create the balanced COPA training set, we manually mirror the original training set by modifying the premise.
Taking the original training set as a starting point, we duplicate the COPA instances and modify their premises so that incorrect alternatives become correct.
Suppose the following original COPA instance:
\copa
{The stain came out of the shirt. What was the CAUSE of this?}
{I bleached the shirt. (Correct)}
{I patched the shirt.}
We create the following balanced COPA instance, where the wrong alternative becomes the correct choice now:
\copa
{\textit{The shirt did not have a hole anymore}. What was the CAUSE of this?}
{I bleached the shirt.}
{I patched the shirt. (Correct)}
This approach is similar to \newcite{niven2019probing}, who create a balanced benchmark of the Argument Reasoning Comprehension Task by negating and rotating its ingredients, exploiting the nature of the task.
However, due to the nature of COPA, we cannot follow their approach and choose to create new premises.

To collect such balanced data, we asked five fluent English speakers who have background knowledge of NLP (see Appendix~\ref{app:guideline} for the detailed guideline). 
Finally, we collected 500 new mirrored instances.
Concatenating it with the original training instances, the balanced COPA consists of 1,000 instances in total.
The corpus is publicly available at \url{https://balanced-copa.github.io}.

\subsection{Quality Evaluation}

To ensure the quality of the mirrored instances, we estimate a human performance using Amazon Mechanical Turk (AMT), a widely-used crowdsourcing platform.
We randomly sample 100 instances from the original COPA training set and 100 instances from the balanced COPA, and asked crowdworkers to solve each instance (see Appendix~\ref{app:amt_form} for an actual screenshot).
%
To avoid noisy workers, we presented our tasks to workers who meet master AMT qualification with at least 10,000 HIT approvals and 99\% HIT approval rate.
Per HIT, we assign three crowd workers and offer 10 cents reward.

\begin{table}[t]
\centering
\begin{tabular}{lrr}
\toprule
Dataset & Accuracy & Fleiss' kappa $k$ \\
\midrule
Original COPA & 100.0   &	0.973 \\
Balanced COPA & 97.0    &	0.798 \\
\bottomrule
\end{tabular}
\caption{Results of human performance evaluation of the original COPA and Balanced COPA.}
\label{tab: amt-evaluation}
\end{table}
From the collected responses, we calculate the accuracy of workers (by majority voting) and inter-annotator agreement by Fleiss' Kappa~\cite{Fleiss81}.
The human evaluation shows that our mirrored instances are comparable in difficulty to the original ones (see Table~\ref{tab: amt-evaluation}).
However, we found that some mirrored instances are a bit tricky at first glance.
But, with a bit more attention, the answer is quite obvious (see Appendix~\ref{app:amt_hard}, for an example).


\section{Experiments}
\label{sec:eval}

\subsection{BERT and RoBERTa on COPA}
\label{sec:bertforcopa}

\begin{table*}[t]
    \centering
    \adjustbox{max width=\textwidth}{
    \begin{tabular}{lllllll}
    \toprule
        Model	    &  Method & Training Data & Overall &   Easy	&   Hard & p-value (\%)	\\
    \midrule
    \newcite{goodwin-etal-2012-utdhlt} & PMI &  unsupervised & 61.8 &     64.7 &     60.0 & 19.8\\
    \newcite{gordon_commonsense_2011-1} & PMI & unsupervised & 	65.4 &     65.8 &     65.2 & 83.5\\
    \newcite{sasaki-etal-2017-handling} & PMI & unsupervised & 	71.4 &     75.3 &     69.0 & 4.8$^\ast$ \\
    Word frequency  & wordfreq  &  COPA & 53.5 & 57.4 &   51.3 & 9.8  \\
    \midrule
    BERT-large-FT   & LM, NSP & COPA &	76.5 ($\pm$ 2.7) &   83.9 ($\pm$ 4.4) &  71.9 ($\pm$ 2.5) & 0.0$^\ast$\\
    RoBERTa-large-FT & LM & COPA & 87.7 ($\pm$ 0.9) &   91.6 ($\pm$ 1.1) &  85.3 ($\pm$ 2.0) & 0.0$^\ast$\\
    
    \bottomrule
    \end{tabular}
    }
    \caption{Model performance on the COPA test set (\emph{Overall}), on \emph{Easy} instances with superficial cues, and on \emph{Hard} instances without superficial cues. p-values according to Approximate Randomization Tests \cite{noreen1989computer}, with $^\ast$ indicating a significant difference between performance on \emph{Easy} and \emph{Hard} $p < 5\%$. Methods are pointwise mutual information (PMI), word frequency provided by the \texttt{wordfreq} package \cite{speer_robyn_2018_1443582}, pretrained language model (LM), and next-sentence prediction (NSP).}
    \label{tab:easy-hard-evaluation}
\end{table*}

In this section we analyze the performance of two recent pretrained language models on COPA: BERT and RoBERTa, an optimized variant of BERT that achieves better performance on the SuperGLUE benchmark \cite{wang2019superglue}, which includes COPA.

We convert COPA instances as follows to make them compatible with the input format required by BERT/RoBERTa.
For a COPA instance $\langle p, a_1, a_2, q \rangle$, where $p$ is a premise, $a_i$ is the $i$-th alternative, and $q$ is a question type (either \emph{effect} or \emph{cause}), we construct BERT's input depending on the question type.
We assume that the first sentence and the second sentence in the next sentence prediction task describe a cause and an effect, respectively.
Specifically, for each $i$-th alternative, we define the following input function:
\begin{eqnarray}
{\rm input}(p, a_i) =
\begin{cases}
\text{\small ``[CLS] $p$ [SEP] $a_i$ [SEP]'' if $q$ is effect} \\
\text{\small ``[CLS] $a_i$ [SEP] $p$ [SEP]'' if $q$ is cause}
\end{cases}
\nonumber
\end{eqnarray}

Part of BERT's training objective includes next sentence prediction. Given a pair of sentences, BERT predicts whether one sentence can be plausibly followed by the other. For this, BERT's input format contains two [SEP] tokens to mark the two sentences and the [CLS] token, which is used as the input representation for next sentence prediction. This part of BERT's architecture makes it a natural fit for COPA.

\begin{table*}[t]
    \begin{minipage}{\textwidth}
        \centering
        \begin{tabular}{llccc}
        \toprule
            Model	&   Training data & Overall &   Easy    &   Hard \\
        \midrule
    	 BERT-large-FT & B-COPA & 74.5 ($\pm$ 0.7) &  74.7 ($\pm$ 0.4) &  \textbf{74.4} ($\pm$ 0.9) \\
    	 BERT-large-FT & B-COPA (50\%) & 74.3 ($\pm$ 2.2) &  76.8 ($\pm$ 1.9) &  72.8 ($\pm$ 3.1) \\
    	 BERT-large-FT & COPA	& \textbf{76.5} ($\pm$ 2.7) &  \textbf{83.9} ($\pm$ 4.4) &  71.9 ($\pm$ 2.5) \\
    	 \midrule
RoBERTa-large-FT & B-COPA    &  \textbf{89.0} ($\pm$ 0.3) &  88.9 ($\pm$ 2.1) &  \textbf{89.0} ($\pm$ 0.8) \\
RoBERTa-large-FT & B-COPA (50\%) &  86.1 ($\pm$ 2.2) &  87.4 ($\pm$ 1.1) &  85.4 ($\pm$ 2.9) \\
RoBERTa-large-FT & COPA       &  87.7 ($\pm$ 0.9) &  \textbf{91.6} ($\pm$ 1.1) &  85.3 ($\pm$ 2.0) \\    	 
        \bottomrule
        \end{tabular}
        \caption{Results of fine-tuned models on Balanced COPA.
        \emph{Easy}: instances with superficial cues, \emph{Hard}: instances without superficial cues.}
        \label{tab:balanced-easy-hard-evaluation}
    \end{minipage}
     \par\bigskip 
\end{table*}

One of the key differences between BERT and RoBERTa is that the next sentence prediction objective is not part of RoBERTa's training objective.
Instead, RoBERTa is trained with masked language modeling only, with its input consisting of multiple concatenated sentences.
To match this training setting, we encode two sentences in a single segment as follows:
\begin{eqnarray}
{\rm input}(p, a_i) =
\begin{cases}
\text{``$<$s$>$ $p$ $a_i$ $<$/s$>$'' if $q$ is effect} \\
\text{``$<$s$>$ $a_i$ $p$ $<$/s$>$'' if $q$ is cause} \\
\end{cases}
\nonumber
\end{eqnarray}

After encoding premise-alternative with BERT or RoBERTa, we take the first hidden representation $\bm{z_i^0}$, i.e.\ the one corresponding to [CLS] or $<$s$>$, in the final model layer and pass it through a linear layer for binary classification:
\begin{eqnarray}
&& y_i = \bm{w}^{\intercal} \bm{z_i^0} + b, \label{eq:score}
\end{eqnarray}
where the parameters $\bm{w} \in \mathbb{R}^{h}$ and $b \in \mathbb{R}$ are learned on the COPA training set.
Finally, we choose the alternative with the higher score, i.e., $a_{\hat{i}}$ with $\hat{i} = \argmax_{i \in \{1, 2\}}y_i$.

For training, we minimize the cross entropy loss with the logits $[y_1; y_2]$ and fine-tune BERT and RoBERTa's parameters.
In our experiments, we use pretrained BERT-large (uncased) with 24 layers, 16 self-attention heads (340M parameters) and pretrained RoBERTa-large with 24 layers, 16 self-attention heads (355M parameters).\footnote{\url{https://huggingface.co/pytorch-transformers/}}
%

\subsection{Training Details}

For training, we consider two configurations: (i) using the original COPA training set (\S\ref{sec:easyhardeval}), and (ii) using B-COPA (\S\ref{sec:balaeval}).
We randomly split the training data into training data and validation data with the ratio of 9:1.
For B-COPA, we make sure that a pair of original instance and its mirrored counterpart always belong to the same split in order to ensure that a model is trained without superficial cues.
For testing, we use all 500 instances from the original COPA test set.

We run each experiment three times with different random seeds and average the results.
We train for 10 epochs and choose the best model based on the validation score.
To reduce GPU RAM usage, we set BERT and RoBERTa's maximum sequence length to 32, which covers all training and test instances. 
We use Adam \cite{kingma2014adam} with warmup, weight decay of 0.01, a batch size of 4, and a gradient accumulation of 8.
We optimize hyperparameters for BERT and RoBERTa separately on the validation set.
For BERT, we test learning rates of 2e-4, 1e-4, 8e-5, 4e-5, 2e-5, and 1e-5, and use warm up proportion of 0.1, with gradient norm clipping of 1.0.
For RoBERTa, we test learning rates of 1e-5, 8e-6, 6e-6, 4e-6, 2e-6, and 1e-6, and use warm up proportion of 0.06, with no gradient norm clipping.


%
\begin{table*}[t]
     \par\bigskip 
    \begin{minipage}{\textwidth}
        \centering
        \begin{tabular}{llccc}
        \toprule
            Model	&   Training data & Overall &   Easy    &   Hard \\
        \midrule
	 BERT-large & B-COPA	& 70.5 ($\pm$ 2.5) &  72.6 ($\pm$ 2.3) &  \textbf{69.1} ($\pm$ 2.7) \\
	 BERT-large & B-COPA (50\%)	& 69.9 ($\pm$ 1.9) &  71.2 ($\pm$ 1.3) &  69.0 ($\pm$ 3.5) \\
	 BERT-large  & COPA	& \textbf{71.7} ($\pm$ 0.5) &  \textbf{80.5} ($\pm$ 0.4) &  66.3 ($\pm$ 0.8) \\
	 \midrule
    RoBERTa-large & B-COPA    &  \textbf{76.7} ($\pm$ 0.8) &  73.3 ($\pm$ 1.5) &  \textbf{78.8} ($\pm$ 2.0) \\
    RoBERTa-large & B-COPA (50\%) &  72.4 ($\pm$ 2.0) &  72.1 ($\pm$ 1.7) &  72.6 ($\pm$ 2.1) \\
    RoBERTa-large & COPA       &  76.4 ($\pm$ 0.7) &  \textbf{79.6} ($\pm$ 1.0) &  74.4 ($\pm$ 1.1) \\
    	 \midrule
     	BERT-base-NSP	& None & \textbf{66.4}	& 66.2	& \textbf{66.7} \\
     	BERT-large-NSP	& None & 65.0	& \textbf{66.9}	& 62.1 \\
        \bottomrule
        \end{tabular}
        \caption{Results of non-fine-tuned models on Balanced COPA.
        \emph{Easy}: instances with superficial cues, \emph{Hard}: instances without superficial cues.}
        \label{tab:balanced-easy-hard-evaluation_nft}
    \end{minipage}
\end{table*}

\subsection{Evaluation on Easy and Hard subsets}
\label{sec:easyhardeval}





To investigate the behaviour of BERT and RoBERTa trained on the original COPA, which contains superficial cues, we split the test set into an \emph{Easy} subset and a \emph{Hard} subset.
The \emph{Easy subset} consists of instances that are correctly solved by the premise-oblivious model described in \S\ref{sec:cues}.
To account for variation between the three runs with different random seeds, we deem an instance correctly classified only if the premise-oblivous model's prediction is correct for all three runs.
This results in the \emph{Easy} subset with 190 instances and the \emph{Hard} subset comprising the remaining 310 instances.
Such an easy/hard split follows similar splits in NLI datasets \cite{gururangan-etal-2018-annotation}.

We then compare BERT and RoBERTa with previous models on the \emph{Easy} and \emph{Hard} subsets.\footnote{For previous models, we use the prediction keys available on \url{http://people.ict.usc.edu/~gordon/copa.html}}
As Table~\ref{tab:easy-hard-evaluation} shows, previous models perform similarly on both subsets, with the exception of \newcite{sasaki-etal-2017-handling}.\footnote{We conjecture that word frequency is another superficial cue exploited by models. To verify this we train a classifier based on word frequencies only \cite{speer_robyn_2018_1443582} and find that this classifier is able to identify the correct alternative better than random chance, but this result is not significant ($p = 9.8\%)$.}
Overall both BERT (76.5\%) and RoBERTa (87.7\%) considerably outperform the best previous model (71.4\%). However, BERT's improvements over previous work can be almost entirely attributed to high accuracy on the \emph{Easy} subset: on this subset, finetuned BERT-large improves 8.6 percent over the model by \cite{sasaki-etal-2017-handling} (83.9\% vs. 75.3\%), but on the \emph{Hard} subset, the improvement is only 2.9 percent (71.9\% vs. 69.0\%).
This indicates that BERT relies on superficial cues.
The difference between accuracy on \emph{Easy} and \emph{Hard} is less pronounced for RoBERTa, but still suggests some reliance on superficial cues.
We speculate that superficial cues in the COPA training set prevented BERT and RoBERTa from focusing on task-related non-superficial cues such as causally related event pairs.

\subsection{Evaluation on Balanced COPA (B-COPA)}
\label{sec:balaeval}

How will BERT and RoBERTa behave when there are no superficial cues in the training set?
To answer this question, we now train BERT and RoBERTa on B-COPA and evaluate on the \emph{Easy} and \emph{Hard} subsets.
The results are shown in Table~\ref{tab:balanced-easy-hard-evaluation}.
The smaller performance gap between \emph{Easy} and \emph{Hard} subsets indicates that training on B-COPA encourages BERT and RoBERTa to rely less on superficial cues.
Moreover, training on B-COPA improves performance on the Hard subset, both when training with all 1000 instances in B-COPA, and when matching the training size of the original COPA (500 instances, \emph{B-COPA 50\%}).
Note that training on \emph{B-COPA 50\%} exposes the model to lexically less diverse training instances than the original COPA due to the high overlap between mirrored alternatives (see \S\ref{sec:balanced-copa}).

These results show that once superficial cues are removed, the models are able to learn the task to a high degree.
This contrasts with \citet{niven2019probing}, who found that BERT's performance on the Argument Reasoning Comprehension Task \cite{habernal-etal-2018-argument} does not exceed random chance level after superficial cues are made uninformative.
A likely explanation for this contrast is the difference in the inherent task difficulties. Argument reasoning comprehension is a high level natural language understanding task requiring world knowledge and complex reasoning skills, while COPA can be largely solved with associative reasoning, as the performance of the PMI-based baselines shows (Table~\ref{tab:easy-hard-evaluation}).
A second possible explanations is BERT's insensitivity to negations \cite{Ettinger2019}. Since \newcite{niven2019probing} made superficial cues uninformative by adding negated instances to the dataset, BERT's insensitivity to negations makes distinguishing between instances and negated instances difficult (see \S\ref{sec:balanced-copa}).

%

%
%
%

%
\subsection{Analysis of sentence pair embeddings}
\label{sec:sentpairanalysis}

The findings presented in the previous sections, namely BERT's and RoBERTa's good performance on COPA in spite of the rather small amount of training data, leads us to the following hypothesis that pretraining enables these models to create an embedding space in which embeddings of plausible sentence pairs are distinguishable from embeddings of less plausible pairs.

To investigate how well the respective embedding spaces of BERT and RoBERTa separate plausible and less-plausible pairs, we train BERT-large and RoBERTa-large \emph{without fine-tuning}.
Specifically, we freeze model weights and train a classifier by parameterizing $\bm{w}$ and $b$ in Equation \ref{eq:score} as a soft-margin Support Vector Machine \cite[SVM,][]{vapnik95}.\footnote{We tune the SVM hyperparameter $C \in \{0.0001, 0.001, 0.01, 0.1, 1\}$ on the validation set.}
We also report results for a simple model that only uses BERT's pretrained next sentence predictor (BERT-base-NSP, BERT-large-NSP), i.e., we choose the alternative with the higher next sentence prediction score.
The results are shown in Table~\ref{tab:balanced-easy-hard-evaluation_nft}.
The relatively high accuracies of BERT-large, RoBERTa-large and BERT-*-NSP show that these pretrained models are already well-equipped to perform this task ``out-of-the-box''.

\subsection{Analysis of sensitivity to cues}
%
%
To analyze the sensitivity of BERT and RoBERTa to superficial cues and to content words, we employ a gradient-based approach, following \cite{Brunner2019}.
Specifically, we define the sensitivity $s_{i,t}$ of the classification score in $i$-th COPA test instance to input token $t$, as follows:
\begin{eqnarray}
    s_{i,t} = \frac{||\bm{g_t}||}{\sum_{t' \in T_i} ||\bm{g_{t'}}||}, \bm{g_t} = \frac{\partial y}{\partial \bm{x_t}},
\end{eqnarray}
where $T_i$ is a sequence of all input tokens in the $i$-th COPA test instance, $y$ is a score function defined by Equation~(\ref{eq:score}), and $\bm{x_t} \in \mathbb{R}^{1024}$ is a position-augmented token embedding of $t$.
We then define the sensitivity $S(k)$ to cue $k$ over all COPA test instances as the average over all $m$ COPA test instances: $S(k) = \frac{1}{m} \sum_i^m s_{i,k}$.

\begin{figure}[t]
\centering
\includegraphics[width=\linewidth]{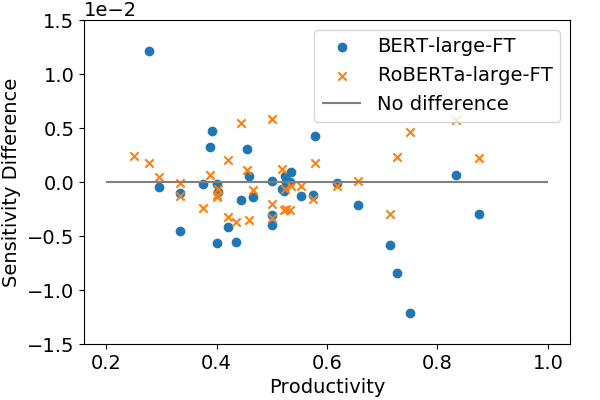}
\caption{Change of sensitivity to superficial cues (in \S 2)
from COPA-trained models to B-COPA-trained models
as a function of their productivity.}
\label{fig:prod_sensitivity}
\end{figure}

We are interested in the change of sensitivity towards cue $t$ of a model trained on original COPA compared to a model trained on Balanced COPA.
We plot this difference as a function of the cue's productivity (Figure~\ref{fig:prod_sensitivity}).
We observe that BERT trained on Balanced COPA is less sensitive to a few highly productive superficial cues than BERT trained on original COPA.
Note the decrease in the sensitivity for cues of productivity from 0.7 to 0.9.
These cues are shown in Table~\ref{tab:sensitivity}.
However, for cues with lower productivity, the picture is less clear, in case of RoBERTa, there are no noticeable trends in the change of sensitivity.



\begin{table}[t]
\centering
\begin{tabular}{lcccc}
\toprule
Cue & $S^{\rm COPA}$ & $S^{\rm B\_COPA}$ & Diff. & Prod. \\
\midrule
   woman & 7.98 & 4.84 & -3.14 &  0.25 \\
  mother & 5.16 & 3.95 & -1.21 &  0.75 \\
    went & 6.00 & 5.15 & -0.85 &  0.73 \\
    down & 5.52 & 4.93 & -0.58 &  0.71 \\
    into & 4.07 & 3.51 & -0.56 &  0.40 \\
\bottomrule
\end{tabular}
\caption{Sensitivity of BERT-large to superficial cues identified in \S\ref{sec:cues} (unit: $10^{-2}$). Cues with top-5 reduction are shown. $S^{\rm COPA}, S^{\rm B\_COPA}$ indicate the mean contributions of BERT-large trained on COPA, and BERT-large trained on B-COPA, respectively.}
\label{tab:sensitivity}
\end{table}



\section{Conclusions}
\label{sec:conclusion}

We established that COPA, an important benchmark of commonsense reasoning, contains superficial cues, specifically single tokens predictive of the correct answer, that allow models to solve the task without actually understanding it.
Our experiments suggest that BERT's good performance on COPA can be explained by its ability to exploit these superficial cues.
BERT performs well on \emph{Easy} instances with such superficial cues, and comparable to previous methods on \emph{Hard} instances without such cues.
RoBERTa, in contrast, represents a real improvement considerably outperforms both BERT and previous methods on \emph{Hard} instances as well.

To allow evaluating models on a benchmark without predictive single tokens, we created the Balanced COPA dataset.
Balanced COPA neutralizes this kind of superficial cue by mirroring instances from the original COPA dataset, thereby removing any differences in token distributions between correct and wrong alternatives. Surprisingly, we found that both BERT and RoBERTa finetuned on Balanced COPA perform comparably overall to the models finetuned on the original COPA.
However, a more detailed analysis revealed quite different behaviour.
Whereas BERT finetuned on original COPA heavily exploited superficial cues, we now find evidence that BERT finetuned on balanced COPA appears to learn some aspects of the task with similar accuracies on both \emph{Easy} and \emph{Hard} instances.
Even more surprisingly, RoBERTa benefits from training on Balanced COPA instances and achieves higher accuracy than on the original COPA with superficial cues.

Two important questions remain unanswered at present, which we plan to explore in future work: 
Even in the presence of superficial cues, RoBERTa does not seem to rely on them.
First, why does RoBERTa not appear to rely on superficial cues, even when they are available?
And second, are the results of our experiments on Balanced COPA specific to BERT and RoBERTa or are all pretrained language models able to exploit superficial cues in COPA and able to solve the task by other means if no such cues are present?



\section*{Acknowledgements}
This work was partially supported by JSPS KAKENHI Grant Number 19K20332 and JST CREST Grant Number JPMJCR1513, including AIP challenge.

\bibliography{emnlp-ijcnlp-2019}

\begin{thebibliography}{29}
\expandafter\ifx\csname natexlab\endcsname\relax\def\natexlab#1{#1}\fi

\bibitem[{Bowman et~al.(2015)Bowman, Angeli, Potts, and
  Manning}]{snli:emnlp2015}
Samuel~R. Bowman, Gabor Angeli, Christopher Potts, and Christopher~D. Manning.
  2015.
\newblock A large annotated corpus for learning natural language inference.
\newblock In \emph{Proceedings of the 2015 Conference on Empirical Methods in
  Natural Language Processing (EMNLP)}. Association for Computational
  Linguistics.

\bibitem[{Brunner et~al.(2019)Brunner, Liu, Pascual, Richter, and
  Wattenhofer}]{Brunner2019}
Gino Brunner, Yang Liu, Dami'an Pascual, Oliver Richter, and Roger Wattenhofer.
  2019.
\newblock On the validity of self-attention as explanation in transformer
  models.
\newblock \emph{ArXiv}, abs/1908.04211.

\bibitem[{Cortes and Vapnik(1995)}]{vapnik95}
Corinna Cortes and Vladimir Vapnik. 1995.
\newblock \href {https://doi.org/10.1023/A:1022627411411} {Support-vector
  networks}.
\newblock \emph{Mach. Learn.}, 20(3):273--297.

\bibitem[{Dasgupta et~al.(2018)Dasgupta, Guo, Stuhlm{\"u}ller, Gershman, and
  Goodman}]{Dasgupta2018EvaluatingCI}
Ishita Dasgupta, Demi Guo, Andreas Stuhlm{\"u}ller, Samuel~J Gershman, and
  Noah~D. Goodman. 2018.
\newblock Evaluating compositionality in sentence embeddings.
\newblock \emph{ArXiv}, abs/1802.04302.

\bibitem[{Devlin et~al.(2019)Devlin, Chang, Lee, and
  Toutanova}]{devlin-etal-2019-bert}
Jacob Devlin, Ming-Wei Chang, Kenton Lee, and Kristina Toutanova. 2019.
\newblock \href {https://doi.org/10.18653/v1/N19-1423} {{BERT}: Pre-training of
  deep bidirectional transformers for language understanding}.
\newblock In \emph{Proceedings of the 2019 Conference of the North {A}merican
  Chapter of the Association for Computational Linguistics: Human Language
  Technologies, Volume 1 (Long and Short Papers)}, pages 4171--4186,
  Minneapolis, Minnesota. Association for Computational Linguistics.

\bibitem[{Ettinger(2019)}]{Ettinger2019}
Allyson Ettinger. 2019.
\newblock \href {http://arxiv.org/abs/1907.13528} {What bert is not: Lessons
  from a new suite of psycholinguistic diagnostics for language models}.

\bibitem[{Fleiss(1981)}]{Fleiss81}
Joseph~L. Fleiss. 1981.
\newblock \emph{{Statistical methods for rates and proportions}}, 2nd edition.
\newblock Wiley, New York.

\bibitem[{Goodwin et~al.(2012)Goodwin, Rink, Roberts, and
  Harabagiu}]{goodwin-etal-2012-utdhlt}
Travis Goodwin, Bryan Rink, Kirk Roberts, and Sanda Harabagiu. 2012.
\newblock \href {https://www.aclweb.org/anthology/S12-1063} {{UTDHLT}:
  {COPACETIC} system for choosing plausible alternatives}.
\newblock In \emph{*{SEM} 2012: The First Joint Conference on Lexical and
  Computational Semantics {--} Volume 1: Proceedings of the main conference and
  the shared task, and Volume 2: Proceedings of the Sixth International
  Workshop on Semantic Evaluation ({S}em{E}val 2012)}, pages 461--466,
  Montr{\'e}al, Canada. Association for Computational Linguistics.

\bibitem[{Gordon et~al.(2011)Gordon, Bejan, and
  Sagae}]{gordon_commonsense_2011-1}
Andrew~S. Gordon, Cosmin~Adrian Bejan, and Kenji Sagae. 2011.
\newblock \href
  {http://ict.usc.edu/pubs/Commonsense%20Causal%20Reasoning%20Using%20Millions%20of%20Personal%20Stories.pdf}
  {Commonsense {Causal} {Reasoning} {Using} {Millions} of {Personal}
  {Stories}}.
\newblock In \emph{25th {Conference} on {Artificial} {Intelligence}
  ({AAAI}-11)}, San Francisco, CA.

\bibitem[{Gururangan et~al.(2018)Gururangan, Swayamdipta, Levy, Schwartz,
  Bowman, and Smith}]{gururangan-etal-2018-annotation}
Suchin Gururangan, Swabha Swayamdipta, Omer Levy, Roy Schwartz, Samuel Bowman,
  and Noah~A. Smith. 2018.
\newblock \href {https://doi.org/10.18653/v1/N18-2017} {Annotation artifacts in
  natural language inference data}.
\newblock In \emph{Proceedings of the 2018 Conference of the North {A}merican
  Chapter of the Association for Computational Linguistics: Human Language
  Technologies, Volume 2 (Short Papers)}, pages 107--112, New Orleans,
  Louisiana. Association for Computational Linguistics.

\bibitem[{Habernal et~al.(2018)Habernal, Wachsmuth, Gurevych, and
  Stein}]{habernal-etal-2018-argument}
Ivan Habernal, Henning Wachsmuth, Iryna Gurevych, and Benno Stein. 2018.
\newblock \href {https://doi.org/10.18653/v1/N18-1175} {The argument reasoning
  comprehension task: Identification and reconstruction of implicit warrants}.
\newblock In \emph{Proceedings of the 2018 Conference of the North {A}merican
  Chapter of the Association for Computational Linguistics: Human Language
  Technologies, Volume 1 (Long Papers)}, pages 1930--1940, New Orleans,
  Louisiana. Association for Computational Linguistics.

\bibitem[{Kingma and Ba(2015)}]{kingma2014adam}
Diederik~P. Kingma and Jimmy Ba. 2015.
\newblock \href {http://arxiv.org/abs/1412.6980} {Adam: {A} method for
  stochastic optimization}.
\newblock In \emph{3rd International Conference on Learning Representations,
  {ICLR} 2015, San Diego, CA, USA, May 7-9, 2015, Conference Track
  Proceedings}.

\bibitem[{Li et~al.(2019)Li, Chen, and Van~Durme}]{li-etal-2019-learning}
Zhongyang Li, Tongfei Chen, and Benjamin Van~Durme. 2019.
\newblock \href {https://www.aclweb.org/anthology/P19-1475} {Learning to rank
  for plausible plausibility}.
\newblock In \emph{Proceedings of the 57th Annual Meeting of the Association
  for Computational Linguistics}, pages 4818--4823, Florence, Italy.
  Association for Computational Linguistics.

\bibitem[{Liu et~al.(2019{\natexlab{a}})Liu, He, Chen, and Gao}]{liu2019multi}
Xiaodong Liu, Pengcheng He, Weizhu Chen, and Jianfeng Gao. 2019{\natexlab{a}}.
\newblock Multi-task deep neural networks for natural language understanding.
\newblock \emph{arXiv preprint arXiv:1901.11504}.

\bibitem[{Liu et~al.(2019{\natexlab{b}})Liu, Ott, Goyal, Du, Joshi, Chen, Levy,
  Lewis, Zettlemoyer, and Stoyanov}]{RoBERTa2019}
Yinhan Liu, Myle Ott, Naman Goyal, Jingfei Du, Mandar Joshi, Danqi Chen, Omer
  Levy, Mike Lewis, Luke Zettlemoyer, and Veselin Stoyanov. 2019{\natexlab{b}}.
\newblock \href {http://arxiv.org/abs/1907.11692} {Roberta: {A} robustly
  optimized {BERT} pretraining approach}.
\newblock \emph{CoRR}, abs/1907.11692.

\bibitem[{Luo et~al.(2016)Luo, Sha, Zhu, Hwang, and
  Wang}]{Luo:2016:CCR:3032027.3032078}
Zhiyi Luo, Yuchen Sha, Kenny~Q. Zhu, Seung-won Hwang, and Zhongyuan Wang. 2016.
\newblock \href {http://dl.acm.org/citation.cfm?id=3032027.3032078}
  {Commonsense causal reasoning between short texts}.
\newblock In \emph{Proceedings of the Fifteenth International Conference on
  Principles of Knowledge Representation and Reasoning}, KR'16, pages 421--430.
  AAAI Press.

\bibitem[{McCoy et~al.(2019)McCoy, Pavlick, and Linzen}]{mccoy2019right}
Tom McCoy, Ellie Pavlick, and Tal Linzen. 2019.
\newblock \href {https://www.aclweb.org/anthology/P19-1334} {Right for the
  wrong reasons: Diagnosing syntactic heuristics in natural language
  inference}.
\newblock In \emph{Proceedings of the 57th Annual Meeting of the Association
  for Computational Linguistics}, pages 3428--3448, Florence, Italy.
  Association for Computational Linguistics.

\bibitem[{Niven and Kao(2019)}]{niven2019probing}
Timothy Niven and Hung-Yu Kao. 2019.
\newblock \href {https://www.aclweb.org/anthology/P19-1459} {Probing neural
  network comprehension of natural language arguments}.
\newblock In \emph{Proceedings of the 57th Annual Meeting of the Association
  for Computational Linguistics}, pages 4658--4664, Florence, Italy.
  Association for Computational Linguistics.

\bibitem[{Noreen(1989)}]{noreen1989computer}
Eric~W Noreen. 1989.
\newblock \emph{Computer-intensive methods for testing hypotheses}.
\newblock Wiley New York.

\bibitem[{Peters et~al.(2018)Peters, Neumann, Iyyer, Gardner, Clark, Lee, and
  Zettlemoyer}]{peters2018deep}
Matthew Peters, Mark Neumann, Mohit Iyyer, Matt Gardner, Christopher Clark,
  Kenton Lee, and Luke Zettlemoyer. 2018.
\newblock \href {https://doi.org/10.18653/v1/N18-1202} {Deep contextualized
  word representations}.
\newblock In \emph{Proceedings of the 2018 Conference of the North {A}merican
  Chapter of the Association for Computational Linguistics: Human Language
  Technologies, Volume 1 (Long Papers)}, pages 2227--2237, New Orleans,
  Louisiana. Association for Computational Linguistics.

\bibitem[{Pfungst(1911)}]{pfungst1911clever}
Oskar Pfungst. 1911.
\newblock \emph{Clever Hans:(the horse of Mr. Von Osten.) a contribution to
  experimental animal and human psychology}.
\newblock Holt, Rinehart and Winston.

\bibitem[{Poliak et~al.(2018)Poliak, Naradowsky, Haldar, Rudinger, and
  Van~Durme}]{poliak-etal-2018-hypothesis}
Adam Poliak, Jason Naradowsky, Aparajita Haldar, Rachel Rudinger, and Benjamin
  Van~Durme. 2018.
\newblock \href {https://doi.org/10.18653/v1/S18-2023} {Hypothesis only
  baselines in natural language inference}.
\newblock In \emph{Proceedings of the Seventh Joint Conference on Lexical and
  Computational Semantics}, pages 180--191, New Orleans, Louisiana. Association
  for Computational Linguistics.

\bibitem[{Roemmele et~al.(2011)Roemmele, Bejan, and
  Gordon}]{roemmele2011choice}
Melissa Roemmele, Cosmin~Adrian Bejan, and Andrew~S Gordon. 2011.
\newblock \href
  {http://people.ict.usc.edu/~gordon/publications/AAAI-SPRING11A.PDF} {Choice
  of plausible alternatives: An evaluation of commonsense causal reasoning}.
\newblock In \emph{AAAI Spring Symposium on Logical Formalizations of
  Commonsense Reasoning}, Stanford University.

\bibitem[{Sap et~al.(2019)Sap, Rashkin, Chen, Bras, and Choi}]{socialiqapaper}
Maarten Sap, Hannah Rashkin, Derek Chen, Ronan~Le Bras, and Yejin Choi. 2019.
\newblock Socialiqa: Commonsense reasoning about social interactions.
\newblock \emph{ArXiv}, abs/1904.09728.

\bibitem[{Sasaki et~al.(2017)Sasaki, Takase, Inoue, Okazaki, and
  Inui}]{sasaki-etal-2017-handling}
Shota Sasaki, Sho Takase, Naoya Inoue, Naoaki Okazaki, and Kentaro Inui. 2017.
\newblock \href {https://www.aclweb.org/anthology/W17-6937} {Handling multiword
  expressions in causality estimation}.
\newblock In \emph{{IWCS} 2017 {---} 12th International Conference on
  Computational Semantics {---} Short papers}.

\bibitem[{Speer et~al.(2018)Speer, Chin, Lin, Jewett, and
  Nathan}]{speer_robyn_2018_1443582}
Robyn Speer, Joshua Chin, Andrew Lin, Sara Jewett, and Lance Nathan. 2018.
\newblock \href {https://doi.org/10.5281/zenodo.1443582}
  {Luminosoinsight/wordfreq: v2.2}.

\bibitem[{Sugawara et~al.(2018)Sugawara, Inui, Sekine, and
  Aizawa}]{sugawara-etal-2018-makes}
Saku Sugawara, Kentaro Inui, Satoshi Sekine, and Akiko Aizawa. 2018.
\newblock \href {https://doi.org/10.18653/v1/D18-1453} {What makes reading
  comprehension questions easier?}
\newblock In \emph{Proceedings of the 2018 Conference on Empirical Methods in
  Natural Language Processing}, pages 4208--4219, Brussels, Belgium.
  Association for Computational Linguistics.

\bibitem[{Wang et~al.(2019)Wang, Pruksachatkun, Nangia, Singh, Michael, Hill,
  Levy, and Bowman}]{wang2019superglue}
Alex Wang, Yada Pruksachatkun, Nikita Nangia, Amanpreet Singh, Julian Michael,
  Felix Hill, Omer Levy, and Samuel~R. Bowman. 2019.
\newblock Super{GLUE}: A stickier benchmark for general-purpose language
  understanding systems.
\newblock \emph{arXiv preprint 1905.00537}.

\bibitem[{Williams et~al.(2018)Williams, Nangia, and
  Bowman}]{williams-etal-2018-broad}
Adina Williams, Nikita Nangia, and Samuel Bowman. 2018.
\newblock \href {https://doi.org/10.18653/v1/N18-1101} {A broad-coverage
  challenge corpus for sentence understanding through inference}.
\newblock In \emph{Proceedings of the 2018 Conference of the North {A}merican
  Chapter of the Association for Computational Linguistics: Human Language
  Technologies, Volume 1 (Long Papers)}, pages 1112--1122, New Orleans,
  Louisiana. Association for Computational Linguistics.

\end{thebibliography}
\bibliographystyle{acl_natbib}

\appendix
\label{appendices}

\section{Balanced COPA New Premise Guidelines}
\label{app:guideline}

We instructed dataset creators with the following guidelines:

\begin{enumerate}
    \item Ensure as much lexical overlap in new premise as the original premise.
    \item Ensure little lexical overlap between premise and alternative, but if a word occurs both in premise and alternatives, it is acceptable to include it in the premise.
    \item Maintain, as much as possible, the length and style between the new premise and the original premise.
    \item Ensure that there is no direct association between the correct alternative and premise. 
    \item Avoid slang.
\end{enumerate}

\section{Amazon Mechanical Turk Form}
\label{app:amt_form}

\begin{figure}[h!]
\centering
\includegraphics[width=\linewidth]{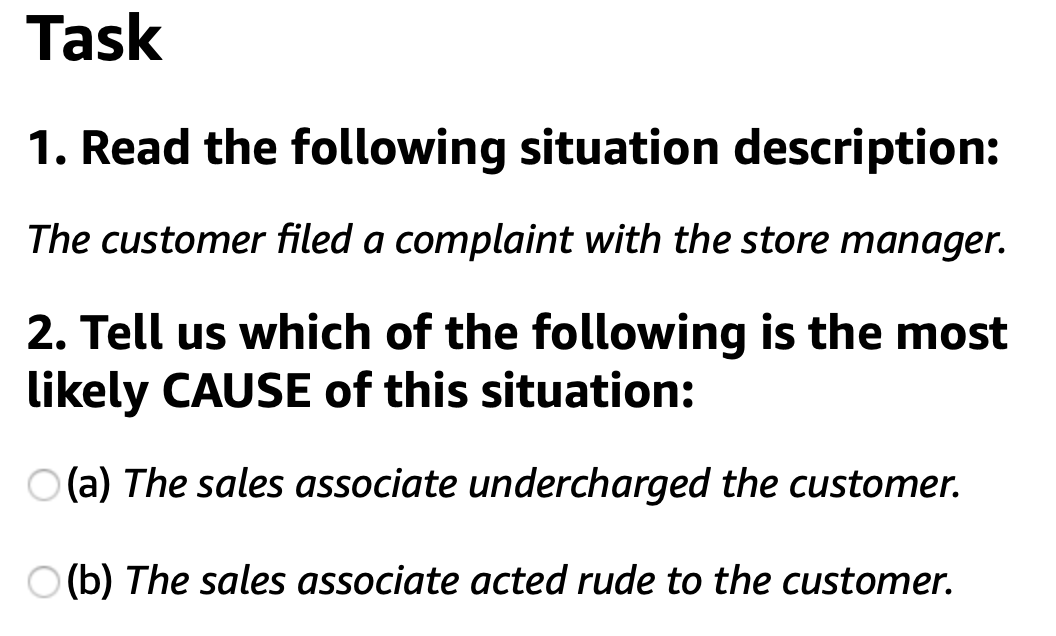}
\caption{Amazon Mechanical Turk task form}
\label{fig:amt-cause-form}
\end{figure}

\section{Example of an instance with low inter-annotator agreement}
\label{app:amt_hard}
\begin{figure}[h!]
    \centering
     \begin{subfigure}[b]{\linewidth}
        \copafig
        {I received a package in the mail. What happened as a result? (effect)}
        {\cmark \tabto{\alternativeindent} The package triggered my curiosity.}
        {\xmark \tabto{\alternativeindent} I took the package to the post office.}
         \caption{Original COPA instance.}
         \label{fig:original-copa2}
     \end{subfigure}
     \par\bigskip 
     \begin{subfigure}[b]{\linewidth}
        \copafig
        {I received \colorbox{green}{someone's} package in the mail. What happened as a result? (effect)}
        {\xmark \tabto{\alternativeindent} The package triggered my curiosity.}
        {\cmark \tabto{\alternativeindent} I took the package to the post office.}
         
        \caption{Mirrored COPA instance.}
        \label{fig:mirrored-copa2}
     \end{subfigure}
     \hfill
    
    \caption{An example of one of mirrored COPA instances with low inter-annotator agreement. Paying attention to the highlighted word is key to picking the correct alternative.}
    \label{fig:tricky-mirrored-instance}
\end{figure}



\end{document}